\pdfoutput=1

\documentclass[11pt]{article}
\usepackage{acl}

\usepackage{times}
\usepackage{latexsym}

\usepackage[T1]{fontenc}

\usepackage[utf8]{inputenc}

\usepackage{microtype}

\usepackage{hyperref}
\usepackage{url}
\usepackage{diagbox}
\usepackage{algorithm}
\usepackage{algpseudocode}
\usepackage{amsmath}
\usepackage{amssymb}
\usepackage{amsmath}
\usepackage{arydshln}
\usepackage{ascii}
\usepackage{bm}
\usepackage{booktabs}
\usepackage{colortbl}
\usepackage{color}
\usepackage{CJK}
\usepackage{dsfont}
\usepackage{enumitem}
\usepackage{forest}
\usepackage{graphics}
\usepackage{graphicx}
\usepackage{latexsym}
\usepackage{multirow}
\usepackage{mwe}
\usepackage{makecell}
\usepackage{microtype}
\usepackage{subfig}
\usepackage{times}
\usepackage{tikz}
\usepackage{tabularx}
\usepackage{url}
\usepackage{wrapfig}
\usepackage{tcolorbox}
\usepackage{xcolor}
\usepackage{adjustbox}

\usepackage{listings}
\usepackage{pifont} 

\definecolor{sred}{RGB}{203, 64, 46}
\definecolor{sblue}{RGB}{44, 73, 135}
\definecolor{sgreen}{RGB}{37, 100, 28}
\definecolor{comments}{RGB}{0, 150, 0}
\definecolor{DarkGreen}{rgb}{0.0, 0.7, 0.0}

\title{LLM$\times$MapReduce-V2: Entropy-Driven Convolutional Test-Time Scaling for Generating Long-Form Articles from Extremely Long Resources}

\author{Haoyu Wang$^{*2}$ \
  Yujia Fu$^{*2}$ \
  Zhu Zhang\thanks{\ \ Equal contribution.}$^{1}$ \
  Shuo Wang$^{\dag1}$ \
  Zirui Ren$^{1}$ \
  Xiaorong Wang$^{3}$ \\
  \textbf{Zhili Li}$^{2}$ \
  \textbf{Chaoqun He}$^{1}$ \
  \textbf{Bo An}$^{4}$ \
  \textbf{Zhiyuan Liu}$^{1}$ \
  \textbf{Maosong Sun}\thanks{\ \ Corresponding authors.}$^{1}$ \\
  $^1$Tsinghua University \\
  $^2$Beijing University of Posts and Telecommunications \\
  $^3$Beijing Jiaotong University \quad
  $^4$Nanyang Technological University \\
  }

\begin{document}

\maketitle
\begin{abstract}
Long-form generation is crucial for a wide range of practical applications, typically categorized into short-to-long and long-to-long generation. While short-to-long generations have received considerable attention, generating long texts from extremely long resources remains relatively underexplored. The primary challenge in long-to-long generation lies in effectively integrating and analyzing relevant information from extensive inputs, which remains difficult for current large language models (LLMs). In this paper, we propose LLM$\times$MapReduce-V2, a novel test-time scaling strategy designed to enhance the ability of LLMs to process extremely long inputs. Drawing inspiration from convolutional neural networks, which iteratively integrate local features into higher-level global representations, LLM$\times$MapReduce-V2 utilizes stacked convolutional scaling layers to progressively expand the understanding of input materials. Both quantitative and qualitative experimental results demonstrate that our approach substantially enhances the ability of LLMs to process long inputs and generate coherent, informative long-form articles, outperforming several representative baselines. \footnote{\ Both LLM$\times$MapReduce-V2 and SurveyEval are publicly available at \url{https://github.com/thunlp/LLMxMapReduce}.}
\end{abstract}

\section{Introduction}

Long-form text generation using large language models (LLMs) holds significant application value and is gaining growing attention~\cite{autosurvey,storm,omnithink}. Based on the amount of information the model should process, long-form text generation can be broadly categorized into two types: {\bf short-to-long} generation and {\bf long-to-long} generation. In short-to-long generation, the model produces long texts from a concise prompt~\cite{fan2019eli5longformquestion, krishna2021hurdlesprogresslongformquestion}.
In contrast, long-to-long generation entails the model producing detailed articles that rely not only on writing prompts but also on a broad range of input data.

\begin{figure}[t]
\centering
\includegraphics[scale=0.350]{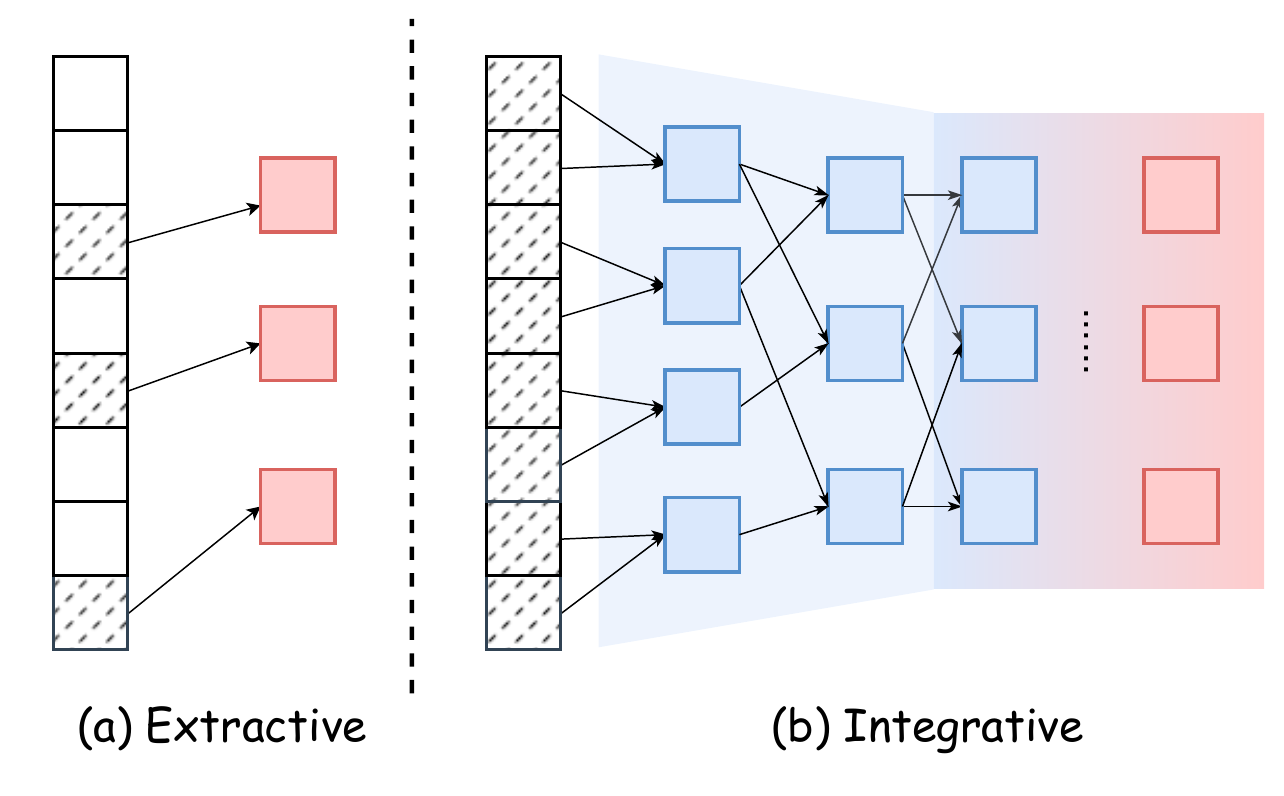}
\caption{Comparison between traditional extractive methods and integrative approach for resource utilization in long-form generation. Extractive methods select relevant content based on queries, which may overlook important information not directly aligned with the query. In contrast, the integrative approach synthesizes a broader range of content, capturing connections for a more comprehensive understanding.}
\label{fig:rag_vs_conv}
\end{figure}

There are two major challenges for long-to-long generation: (1) {\bf resource collection}: retrieving relevant materials for the given topic; and (2) {\bf resource utilization}: effectively integrating these materials to produce informative and cohesive results. Several recent studies focus on improving the resource collection process. For example, STORM~\citep{storm} uses a multi-agent system to pose questions from various perspectives, thereby expanding the coverage of retrieved documents. OmniThink~\citep{omnithink} further develops a growing information tree to progressively expand and deepen the knowledge scope of the collected resources. In real-world scenarios, the relevant resources can be vast~\citep{autosurvey}, making it challenging for modern LLMs or even human experts to extract and synthesize key insights from large volumes of information while analyzing and identifying significant patterns.

Therefore, we focus on enhancing the \textbf{resource utilization} capabilities of LLM-based \textbf{long-to-long} generation frameworks. 
To address the issue that the collected resources exceed the effective context length of LLMs, most existing methods employ extractive techniques to compress the resources~\citep{autosurvey,omnithink}. A common approach is to use embedding models to identify the most relevant chunks based on the queries. 
A major limitation of extractive methods is that they may overlook important content that, while relevant, does not directly align with the given queries. This can include critical analyses, nuanced insights, or broader contextual information that might not be immediately similar but could provoke deeper reflection or contribute to a more comprehensive understanding of the topic.

In this work, we shift from traditional {\bf extractive} methods to {\bf integrative} approaches, aiming to synthesize a broader range of information and draw connections between different pieces of content to create a more holistic and nuanced representation. 
Specifically, we begin with a theoretical analysis of the long-to-long generation task from the information bottleneck perspective.
This analysis underscores the importance of intermediate textual representations, for which we introduce a skeleton and a series of resource digests. By enhancing the informativeness of these intermediate elements, we can theoretically improve the lower bound on the amount of information in the final output.

To facilitate effective information aggregation, we propose a novel randomized convolutional test-time scaling method.  Our approach draws inspiration from the classic convolutional neural network~\citep{lecun1998gradient}, which progressively abstracts local features into high-level global representations, a technique widely used in image processing. We also introduce an information entropy estimation module to guide the convolution process, helping the test-time scaling procedure consistently enhance the informativeness of the results. The resulting long-to-long generation framework, which we term {\em LLM$\times$MapReduce-V2}, effectively helps existing LLMs process extremely long sequences.

Moreover, to evaluate the performance of the proposed integrative framework in comparison to previous extractive methods, we develop a high-quality survey writing benchmark, {\em SurveyEval}. This benchmark consists of academic surveys covering diverse topics, along with their corresponding reference papers. To the best of our knowledge, SurveyEval is the first scalable evaluation benchmark that includes surveys paired with complete reference papers. We selected the survey writing task because it is a quintessential example of generating articles from extensive resources. This task requires the model to thoroughly comprehend the provided reference papers and synthesize informative results that reflect both the current state and future trends of a specific topic. Experimental results on SurveyEval demonstrate that our proposed method consistently outperforms several representative baselines, showcasing the effectiveness of the proposed integrative method.

Our main contributions include:
\begin{itemize}
    \item We conduct a theoretical analysis of the \textbf{long-to-long} generation task, identifying that the key challenge lies in constructing and leveraging informative intermediate representations.
    \item We create a high-quality \textbf{long-to-long} generation benchmark \textbf{SurveyEval}, the first evaluation benchmark in the domain of computer science that pairs surveys with complete reference papers, enabling a thorough comparison of \textbf{resource utilization} capabilities. 
    \item We propose an entropy-driven convolutional test-time scaling framework \textbf{LLM$\times$MapReduce-V2} to use \textbf{integrative} method to solve the \textbf{resource utilization} problem in the \textbf{long-to-long} scenario, with at least 32.9\% improvement in the reference utilization rate and better than the extractive baseline in other dimensions. 
\end{itemize}

\section{Information Bottleneck Analysis}
Long-to-long generation necessitates information compression to conform to the resources within the context window of LLMs and depends on the intermediate representation for constructing the final output, which aligns with the Information Bottleneck (IB)~\citep{tishby2015deeplearninginformationbottleneck} theory. 
It has the following basic forms:
\begin{equation}
    IB(X,Y)=I(Z;Y)-\beta I(X;Z),
    \label{eq:base}
\end{equation}
where $X$ is the input source, $Z$ is the intermediate representation and 
$Y$ is the output. 
$I(\cdot , \cdot)$ represents the mutual information between them. $\beta$ denotes a positive Lagrange multiplier. 

%
Let $X$ be the input materials, which include the topic $T$ of the required output article (i.e., $Y$) and the provided resources $R$, which may be very lengthy. For intermediate representations, we introduce the skeleton $S$, aligned with the output $Y$, and the digests $D$, which are compressed summaries derived from the resources $R$. The information bottleneck can be given by
\begin{equation}
    IB(X,Y) = I(Y;D) - \beta H(D),
    \label{eq:new_form}
\end{equation}
where $H(\cdot)$ represent the information entropy. The detailed derivation from Eq.~(\ref{eq:base}) to Eq.~(\ref{eq:new_form}) can be found in Appendix~\ref{sec:ib_appendix}.

Subsequently, given the information inclusion relationship between the variables, we can get the upper and lower bounds of IB:
\begin{equation}
\begin{split}
    IB(X,Y) \ge & \mathrm{min}((1-\beta)H(D) - H(D|Y), \\
    &H(S)-\beta H(D)), \\ 
    IB(X,Y) \le & H(Y|D) + (1-\beta)H(D). 
\end{split}
\label{eq:bound}
\end{equation}

The detailed derivation process can also be found in Appendix~\ref{sec:ib_appendix}. The bounds shown in Eq.~(\ref{eq:bound}) imply four optimization objectives for the long-to-long generation task:
\begin{itemize}
    \item Maximizing $(1-\beta)H(D)$, which means improving the information in the digests.
    \item Maximizing $H(S)$, which means enhancing the information in the skeleton.
    \item Minimizing $H(D|Y)$, which means reducing the information in the digests that are not used in the final output.
    \item Maximizing $H(Y|D)$, which means incorporating additional information beyond the digest when writing the Survey.
\end{itemize}

In this work, we focus on optimizing the first three objectives to improve the lower bound of the information bottleneck. Optimization of the last objective is left for future work.



\section{LLM\texorpdfstring{$\times$}{x}MapReduce-V2}

Guided by the IB principle, our method employs skeleton-guided digest generation to more effectively extract information from full papers (Sec.~\ref{sec:digest}), entropy-driven convolution and a best-of-N self-refinement mechanism to enhance skeleton quality (Sec.~\ref{sec:outline}), and topology-aware content generation to leverage the information in the digests (Sec.~\ref{subsec:write}).




\subsection{Initialization}
\label{subsec:init}

\paragraph{Survey Tree Construction} 
We employ the idea of the structured information protocol \citep{zhou2024llmtimesmapreducesimplifiedlongsequenceprocessing} to effectively compress the provided materials and fully utilise them.
Throughout the process, both the skeleton and paper digests are parsed into a tree structure that mirrors the generated markdown document. We denote this tree as $\mathcal{T} = (V, E)$, where $V$ is the set of nodes corresponding to section headings, and $E$ defines the parent-child relationships between these nodes. Each skeleton node consists of two key components: Digest Construction, which outlines how to build paper digest nodes, and Digest Analysis, which specifies how these digest nodes will be utilized during the writing process. Figure~\ref{fig:section-structure} illustrates an example of the skeleton structure.


\begin{figure}[h]
\begin{tcolorbox}[colback=blue!2,colframe=blue!50!black]
\small
\textcolor{sblue}{\textbf{\#\#\# 2.1 Section Title}} \\
\textcolor{sred}{\textbf{Digest Construction:}} \\
\texttt{Write about what information should be extracted from the full paper in this section.} \\
\textcolor{sgreen}{\textbf{Digest Analysis:}} \\
\texttt{Write about how to organize and analyse papers ["BIBKEY1", "BIBKEY2"] with executable step.}
\end{tcolorbox}
\caption{Example of the structure in the skeleton.}
\label{fig:section-structure}
\end{figure}



\begin{figure*}[t]
\hspace{0.25cm} 
\includegraphics[scale=0.70]{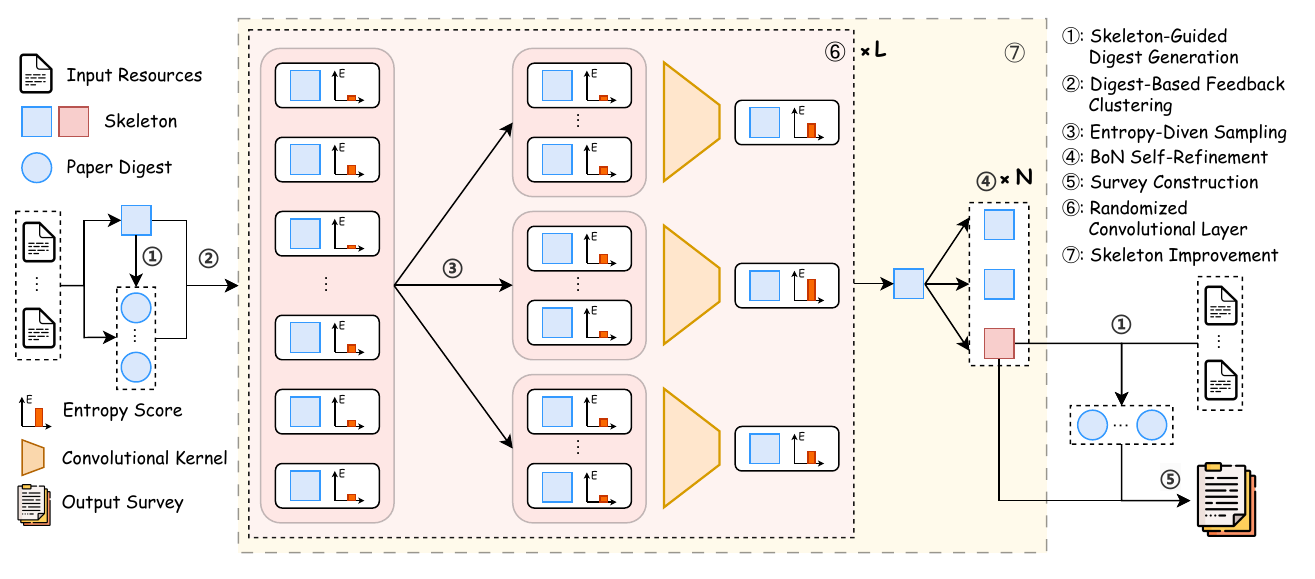}
\caption{The pipeline of LLM$\times$MapReduce-V2. LLM$\times$MapReduce-V2 can be roughly divided into three stages. In the Initialization phase, LLM$\times$MapReduce-V2 initializes the skeleton based on the vast resources and the given topic, and generates the corresponding structured digests. In the Skeleton Improvement phase, LLM$\times$MapReduce-V2 utilizes the feedback from the digests to refine the skeleton, which is guided by entropy-driven random sampling and multi-layer convolution for feedback aggregation. Additionally, a series of Best-of-N iterations are employed to further enhance the skeleton. In the Survey Construction phase, LLM$\times$MapReduce-V2 regenerates structured digests based on the optimized skeleton and performs topology-aware content generation to produce the final survey.}\label{fig:LLMxMapReduce-V2}
\end{figure*}

\paragraph{Skeleton Initialization}
Before generating the digest, an initial skeleton framework should be established based on the given topic \(T\) and a collection of reference resources \(R = \{r_1, r_2, \ldots, r_K\}\). To balance efficiency and performance, the references are first grouped into clusters, denoted by \(\mathcal{C}(R) = \{C_1, C_2, \ldots, C_J\}\), such that \(\bigcup_{j=1}^J C_j = R\). For each cluster \(C_j\), a local skeleton is generated using an LLM-based initialization function \(\mathcal{I}(\cdot)\), and then aggregated using the LLM-based function \(f_{\text{agg}}\) to form a unified initial skeleton:
\begin{equation}
S^{(0)} = f_{\text{agg}}(\sum_{j=1}^J S_j) = f_{\text{agg}}(\sum_{j=1}^J\mathcal{I}(T, C_j)).
\nonumber
\end{equation}

\paragraph{Skeleton-Guided Digest Generation}
\label{sec:digest}
To more accurately and comprehensively compress the content of each reference, the skeleton is used to guide the digest generation. As shown in Figure~\ref{fig:section-structure}, the skeleton includes a Digest Construction component that directs the creation of the digests. Based on the general guidelines provided by the skeleton and the specific content of each reference article $r$, the LLM generates a concise digest $D_r$ tailored to the current skeleton. Furthermore, to foster collaborative optimization between the skeleton and the digests, we require the LLM to propose associated feedback $F_r$ for the skeleton, which provides informative suggestions for the subsequent skeleton improvement process.



\subsection{Skeleton Improvement}
\label{sec:outline}
The skeleton plays a pivotal role in bridging the input and output. Its Digest Construction component guides the extraction of information from references into the digest, while the Digest Analysis part provides instructions for organizing the digests into the final survey content. To fully leverage the potential of test-time scaling and obtain better skeletons, we design two mechanisms: Entropy-Driven Convolution and Best-of-N Self-Refinement.

Inspired by residual~\citep{he2015deepresiduallearningimage}, where $H(x) = x + f(x)$, we develop feedback $\Delta S$ to modify the skeleton, rather than directly generating a new one. This approach better captures the differences between intermediate skeletons, reducing information redundancy for LLMs during the process. Each feedback $\Delta S$ first modifies the base skeleton to produce the updated version $S + \Delta S$, after which the information entropy is evaluated. This entropy is then used to guide the improvement of the skeleton.

To better quantify the information entropy within the skeleton, we split it into two parts: the title structural information entropy \(H_T(S)\) and the chapter description information entropy \(H_C(S)\). Their combined effect is modelled as
\begin{equation}
H(S) = H_T(S) + H_C(S).
\nonumber
\end{equation}

We use LLM-as-judge~\citep{surveyllmasajudge} to get a score out of ten as an estimation of information entropy. 

\subsubsection{Entropy-Driven Convolution}
\label{subsec:conv}

\paragraph{Digest-Based Feedback Clustering}
Based on the initialized skeleton $S^{(0)}$, we have generated new digests $D_r$ and feedback $F_r$. During this process, we need to aggregate the information within each cluster $C_j$ to generate the initial skeleton modification suggestions at the cluster level. Specifically, for cluster $C_j$, we use an LLM-based function $f_{\text{part}}$ to aggregate the information within it and generate partial feedback:
\begin{equation}
\Delta S_j^{(0)} = f_{\text{part}}(\bigoplus _{r \in C_j} D_r, \bigoplus_{r \in C_j} F_r),
\nonumber
\end{equation}
where $1 \le j \le J$, and $\Delta S_j^{(0)}$ represents the modification feedback based on the information within $C_j$. All initial partial feedback will enter multiple randomized convolutional layers for further aggregation.



\paragraph{Entropy-Driven Sampling and Convolution}
Inspired by the hierarchical feature aggregation in convolutional neural networks, we perform multi-layer convolution on the aggregated partial skeleton feedback. Because of the absence of natural spatial adjacencies between different digests, we incorporate an entropy-driven randomized sampling process. At the \(l\)-th layer, each feedback item \(\Delta S_i^l\) is sampled with a probability defined by:
\begin{equation}
p^{(l)}(\Delta S_i^{(l)}) = \frac{H(S+\Delta S_i^{(l)})}{\sum_{i=0}^{N} H(S + \Delta S_i^{(l)})},
\nonumber
\end{equation}
where $N$ is the number of feedback in this layer. From this distribution, multiple sets of feedback items are selected:
\begin{equation}
\Delta \hat{S}^{(l)}_{j} = \text{Sample}\Big(\{\Delta S_i^{(l)}\},\, p^{(l)},\, k\Big).
\nonumber
\end{equation}
The number of sets $k$ is determined by hyperparameters \texttt{result num}, i.e., $ 1 \leq k \leq  \texttt{result num}$. These sampled feedback sets are then integrated parallelly using  \(f_{\text{conv}}\) as a convolution function:
\begin{equation}
\Delta S^{(l+1)}_{j} = f_{\text{conv}}\Big(\Delta \hat{S}^{(l)}_{j} \Big), 
\nonumber
\end{equation}
where $ 1 \leq l \leq  L $. And we select \texttt{top-k} feedback into the next layer. After \(L\) layers, the refined skeleton is obtained by selecting the best one of the last layer:
\begin{align}
& S_{\text{refine}} = S + \arg\max_{\Delta S^{L}_{j}} H(S + \Delta S^{L}_{j}) .
\nonumber
\end{align}

\subsubsection{Best-of-N Self-Refinement}
\label{subsec:bon}
After modifying by digest-based feedback, we use the Best-of-N strategy to make overall adjustments and organization. Specifically, \texttt{best-of} candidate feedbacks are independently generated from the $S_{\text{refine}}$, and the one with the highest entropy is selected:
\begin{align}
& S^{c+1} = S^{c} + \arg\max_{\Delta S_{i}^{c}} H(S^{c} + \Delta S_{i}^{c}),
\nonumber
\end{align}
where $ 1 \leq i \leq  \texttt{best-of}$. This will repeat \texttt{self-refinement step} times, i.e., $ 1 \leq c \leq  \texttt{self-refinement step} $. This selection ensures that the final skeleton \(S^*\) exhibits superior global information integration beyond references.



\subsection{Topology-Aware Content Generation} 
\label{subsec:write}
In the final stage, the optimized skeleton \(S^*\) and the corresponding digests \(\{D_r^*\}\) are used to generate the final survey. Because both the skeleton and the digests adhere to the tree structure \(\mathcal{T} = (V, E)\), each node \(v \in V\) corresponds to a section of the survey. We generate each section's content in node-level to reduce the number of details for fully utilizing the information in digests.

The content for each leaf section is generated using a function \(g_{\mathrm{leaf}}(\cdot)\), which is more focused on the utilization of details and comparison between specific works in multiple digests:
\begin{equation}
y_v = g_{\mathrm{leaf}}\Big(s_v^*, \{d_{r,v}^*\}_{r \in R}\Big),
\nonumber
\end{equation}
where \(s_v^*\) represents the refined skeleton Digest Analysis part for node \(v\) and \(d_{r,v}^*\) is the digest information from reference \(r\) for that section. 

As for the non-leaf section, to make the parent chapter more overarching and comprehensive, sub-section contents are additionally introduced in \(g_{\mathrm{non-leaf}}(\cdot)\):
\begin{equation}
y_v = g_{\mathrm{non-leaf}}\Big(s_v^*, \{d_{r,v}^*\}_{r \in R}, \{y_{v'}\}_{e_{v\rightarrow v'} \in E}\Big).
\nonumber
\end{equation}



\section{Experiment}
\subsection{Dataset}
\label{subsec:dataset}
We have developed a high-quality survey writing benchmark, {\bf SurveyEval}, to support our experimental framework. To the best of our knowledge, SurveyEval is the first evaluation benchmark in the domain of computer science that pairs surveys with complete reference papers. In total, we collected 384 survey papers from the Internet, which together cite over 26,000 references.

Given that running, evaluating, and manually assessing the algorithms is time-consuming and labour-intensive, and to align with the AutoSurvey topic number (i.e., 20 surveys), we selected 20 articles from this collection as the test set. Detailed information on dataset construction and metadata is provided in Appendix~\ref{sec:Detail_Test_Surveys}.



\subsection{Baselines}
\label{subsec:baselines}
We evaluate LLM$\times$MapReduce-V2 against three baselines, all powered by Gemini-2.0-flash-thinking-exp-1219~\citep{geminiteam2024geminifamilyhighlycapable}. The input to each baseline consists of the title and full reference papers from the test set. The baselines include
\begin{itemize}
    \item {\em Vanilla}: Directly feeding the topic and full text of all referenced articles into the model for inference via standard decoding. 
    \item {\em Vanilla+Skeleton}: Explicitly generating a skeleton before writing the full output, inspired by the AgentWrite framework~\citep{bai2025longwriter}.
    \item {\em AutoSurvey}~\citep{autosurvey}: A RAG-based academic survey generation framework. We applied the settings and parameters reported in their original work. 
\end{itemize}





The implementation details can be found in Appendix \ref{sec:baseline_detail}.

\subsection{Evaluation Metrics}
\label{subsec:metrics}
\subsubsection{Automatic Metrics} 
The metrics are grouped into four main dimensions, with scores ranging from 0 to 100.

\paragraph{Structure-Oriented Metric}
This metric is used to evaluate the logical organization and coherence of each section, strictly adhering to the structural criteria of AutoSurvey. Details can be found in Appendix~\ref{sec:structure_detail}.

\paragraph{Content-Oriented Metrics}
\label{sec:SurveyEval Quality}
The evaluation metrics for assessing content quality are briefly introduced below. For a detailed explanation, please refer to Appendix~\ref{sec:content_detail}.
\begin{itemize}
    \item {\em Faithfulness}: The precision of sentences with citations in the final output, where correctness is measured by whether the sentence is accurately supported by the cited resources (i.e., the reference papers in the survey writing task).  
    \item {\em Relevance}: The degree to which the content aligns with the research topic, assessing how well the content stays focused on the required research question.
    \item {\em Language}: The assessment of academic formality, clarity, and the avoidance of redundancy in the survey. This metric evaluates the overall quality of writing, ensuring the language is clear, concise, and appropriate for an academic audience.
    \item {\em Criticalness}: The extent to which the survey demonstrates critical analysis, provides original insights and identifies future research directions. This metric evaluates how well the survey goes beyond summarizing existing work, offering thoughtful critiques and highlighting gaps, challenges, or opportunities for further investigation.
\end{itemize}
\begin{table*}[t]
\centering
\begin{tabular}{l|rrrrrrrrr}
    \toprule
    \multirow{2}{*}{\bf Methods} & \multirow{2}{*}{\bf Struct.} &\multicolumn{4}{c}{{\bf Content}} & \multicolumn{2}{c}{{\bf Claim}} & \multicolumn{2}{c}{{\bf Reference}} \\
    \cmidrule(lr){3-6} \cmidrule(lr){7-8} \cmidrule(lr){9-10}
    &  & Fait. & Rele. & Lang. & Crit.  & Num. & Dens. & Prec. & Recall  \\
    \midrule
    \multicolumn{10}{c}{\em Standard Decoding} \\
    \midrule
    Vanilla & 94.44 & 96.43 & \bf 100.00 & \bf 96.50 & 37.11  & 78.75 & \bf 74.64  & 25.48 & 26.46 \\
    \quad + Skeleton & \bf 98.95 & \bf 97.03 & \bf 100.00 & 95.95 & \bf 41.01  & \bf 135.15 & 72.96 & \bf 62.60 & \bf 65.11  \\
    \midrule
    \multicolumn{10}{c}{\em Test-Time Scaling} \\
    \midrule
    AutoSurvey & 86.00  & 93.10 & \bf 100.00 & 92.90 & 68.39  & 423.35 & 31.97 & 50.12 & 51.73  \\
    LLM$\times$MR-V2 & \bf 95.00 & \bf 97.22 & \bf 100.00 & \bf 94.34 & \bf 71.99  & \bf 474.90 & \bf 52.23  & \bf 95.50 & \bf 95.80 \\
    \bottomrule
\end{tabular}
\caption{Performance of the methods evaluated on SurveyEval. For details on the evaluation dimensions, please refer to Section~\ref{subsec:metrics}. The highest scores within each category are bolded.
}
\label{tab:main_result}
\end{table*}
\paragraph{Claim-Oriented Metrics}  
To assess the information amount and density of the survey, we drew inspiration from FactScore~\cite{factscore} to extract claims from the surveys, ensuring that duplicates were removed. Based on this approach, we designed the following two metrics to evaluate both the richness and compactness of the information presented. The full extraction and deduplication procedures are detailed in Appendix~\ref{sec:SurveyEval CD}.  
\begin{itemize}  
   \item {\em Number of Claims}: The total count of unique and accurate claims identified within the text. This metric evaluates the breadth of information presented in the survey by counting the number of distinct, informative claims made. 
   \item {\em Density of Claims}: The ratio of unique claims to the total number of extracted claims before deduplication. This metric reflects the concentration of distinct, relevant information within the survey, indicating how efficiently the content conveys valuable insights. A higher density suggests a more focused and information-rich survey, whereas a lower density may imply redundancy or irrelevant content.

\end{itemize}

\paragraph{Reference-Oriented Metrics} To assess the effective utilization of the provided references in the generated survey, we propose two metrics that measure the coverage and inclusion of references. These metrics aim to quantify the extent to which the input references contribute to the final content, ensuring both precision and comprehensiveness in reference usage. Specifically, we define the following metrics. Detailed can be found in Appendix~\ref{sec:Reference Evaluation Details}
\begin{itemize}
\item {\em Precision}: This metric quantifies the proportion of the input references that are correctly cited at least once in the survey. Precision evaluates how well the references are incorporated into the survey, ensuring that each reference is appropriately acknowledged in the text. A higher precision score indicates that most or all of the provided references have been correctly used in the survey, reflecting thorough integration of the source material.
\item {\em Recall}: Recall measures the total number of input references that appear at least once in the generated survey. This metric captures the breadth of reference inclusion, providing an indication of how many of the input references were utilized overall. A higher recall suggests a more comprehensive survey, where a larger proportion of the input references are cited, while a lower recall may indicate that some references were overlooked or underutilized.
\end{itemize}

These two metrics together provide a balanced assessment of reference use in the survey, with precision focusing on the correct application of references and recall emphasizing their overall inclusion. Both are crucial for ensuring that the survey is grounded in relevant prior work while also reflecting an efficient use of the provided references.

\begin{figure}[h]
\centering
\includegraphics[scale=0.30]{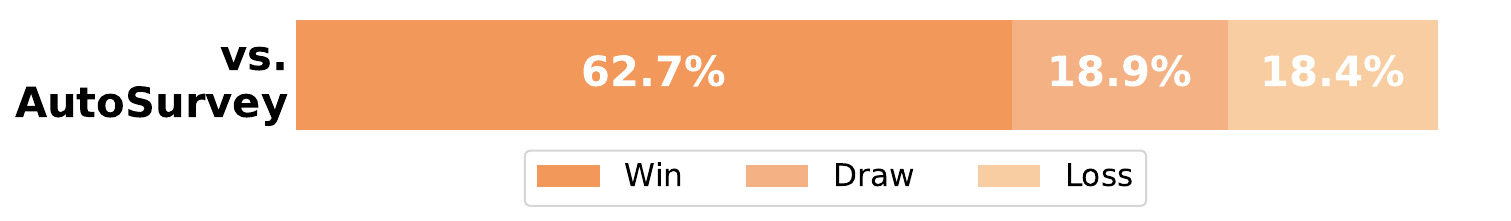}
\caption{Human-evaluated win rate of LLM$\times$MapReduce-V2 compared to AutoSurvey on the test set.}
\label{fig:win_rate}
\end{figure}

\subsubsection{Human Evaluation}
To enable a more reliable comparison of the overall quality between LLM$\times$MapReduce-V2 and other baselines, we conduct a human evaluation. In this process, assessors are asked to determine which survey performs better on the same topic. The {\em win rate} is then computed based on these comparisons. Figure~\ref{fig:win_rate} shows the evaluation results. Further details of the evaluation procedure can be found in Appendix~\ref{sec:Human Evaluation Details}.


\begin{figure*}[ht] 
    \centering 
    \subfloat[Effect of Convolutional Layer]{
        \includegraphics[width=0.3\textwidth]{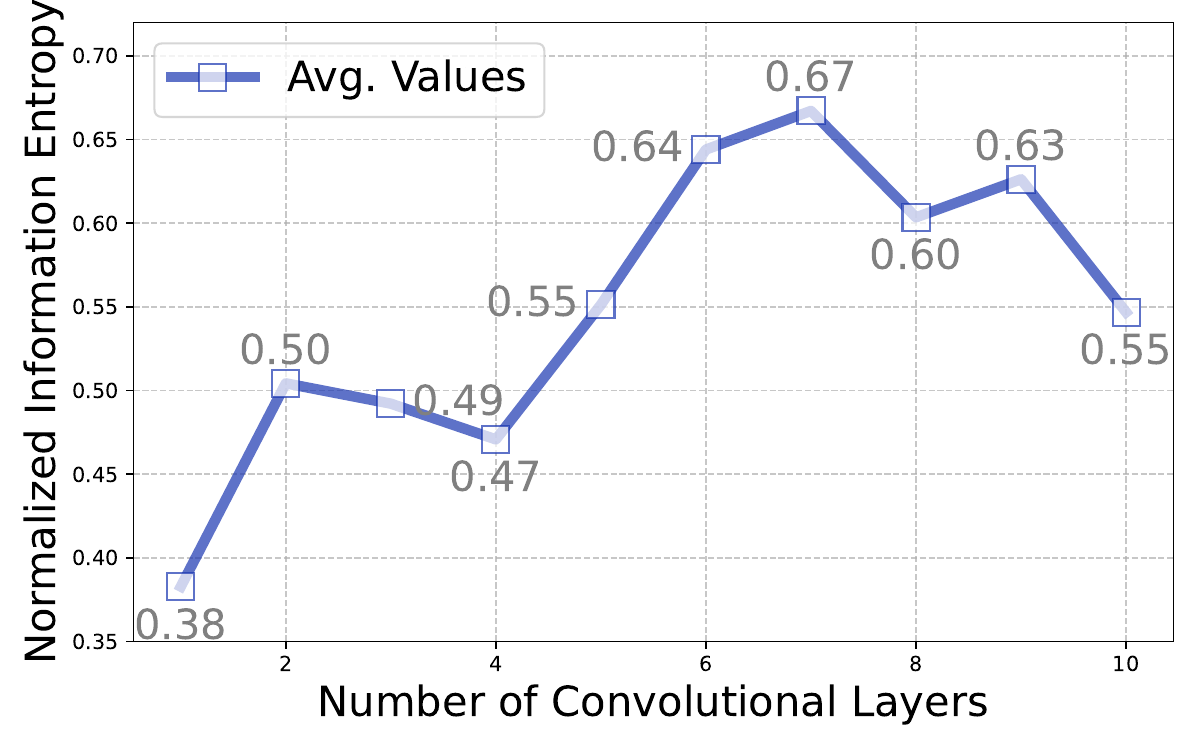}
        \label{fig:conv}
    }
    \subfloat[Effect of Convolutional Kernel]{
        \includegraphics[width=0.3\textwidth]{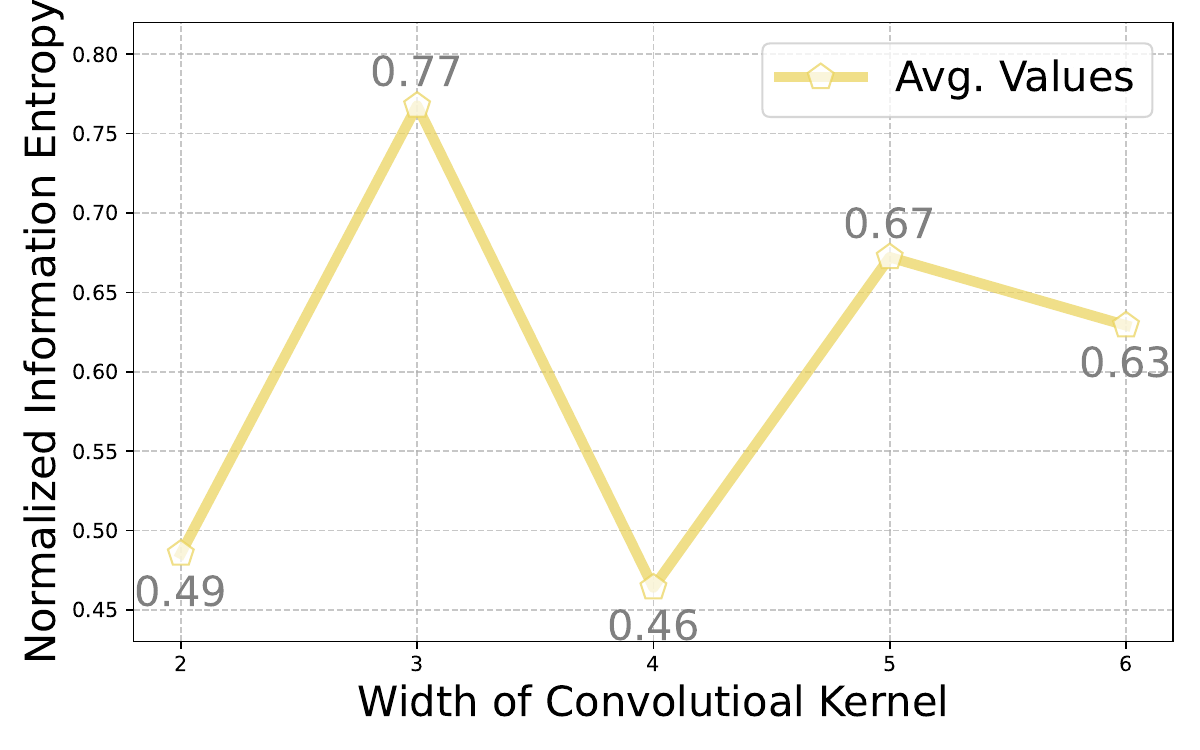}
        \label{fig:recep}
    }
    \subfloat[Effect of Self-Refinement]{
        \includegraphics[width=0.3\textwidth]{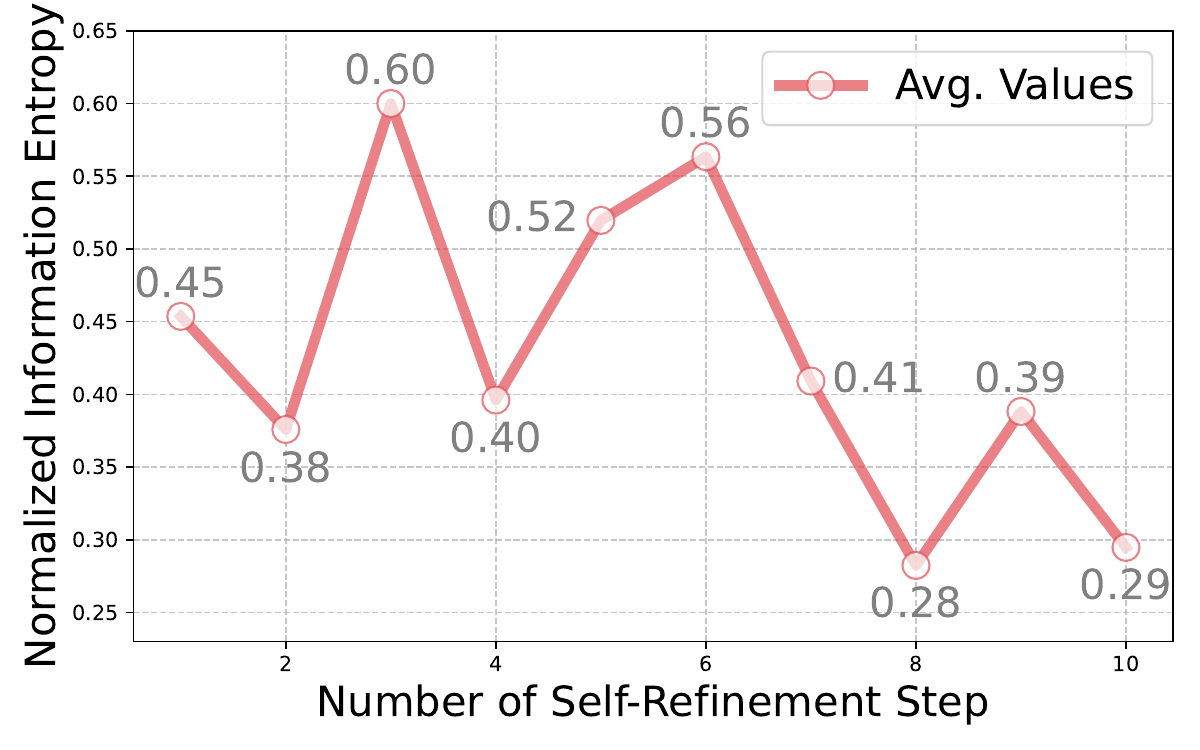}
        \label{fig:self}
    }
    \caption{Analysis of the components in LLM$\times$MapReduce-V2. We use the normalized information entropy score as the evaluation metric for the skeleton, which reflects the informativeness of the intermediate results.}
    \label{fig:three-images}
\end{figure*}

\subsection{Main Results}

Table \ref{tab:main_result} presents the results of four involved methods across four dimensions. The results highlight that LLM$\times$MapReduce-V2 consistently outperforms the baseline methods in most dimensions.

LLM$\times$MapReduce-V2 achieves a score of 95.00 in terms of structural metrics, which is higher than AutoSurvey (i.e., 86.00). The content-oriented metrics, which are crucial for understanding the effectiveness of the methods in generating meaningful and relevant output, show a significant advantage for LLM$\times$MapReduce-V2. In terms of the faithfulness, LLM$\times$MapReduce-V2 scores 97.22, outperforming AutoSurvey (i.e., 93.10). LLM$\times$MapReduce-V2 also performs very well in critical thinking, with a score of 71.99, better than that of AutoSurvey (i.e., 68.39) and those of the standard decoding baselines.

When evaluating the claims, LLM$\times$MapReduce-V2 generates the largest number of informative claims, highlighting the effectiveness of the proposed entropy-driven convolutional test-time scaling mechanism. Additionally, LLM$\times$MapReduce-V2 exhibits a significantly higher density than AutoSurvey, underscoring the superiority of the integrative approach over the extractive method. Although standard decoding strategies can achieve a high claim density, the total number of unique claims is significantly lower than that of test-time scaling approaches.

Finally, LLM$\times$MapReduce-V2 outperforms the baselines in the reference metrics as well, achieving the highest precision (i.e., 95.50) and recall (i.e., 95.80), significantly surpassing both standard decoding baselines and AutoSurvey. These results demonstrate that LLM$\times$MapReduce-V2 excels at leveraging extensive references, offering a substantial advantage in tasks that require advanced information integration across large-scale resources.

\subsection{Analysis of the Components}
\label{subsec:hyperparameter_search}
The skeleton serves as a pivotal component, acting as a bridge between the digest construction and the final output content. Due to its critical role, it demands more computational resources for refinement. We have devised two mechanisms for harnessing test-time scaling, namely Entropy-Driven Convolution and Best-of-N Self-Refinement, with the aim of achieving the desired enhancement. 
In this section, we will delve into these two modules from the information entropy perspective to analyse the performance under different settings.



\paragraph{Entropy-Driven Convolution}
In this module, we focus on the Convolutional Layer and the Width of the Convolutional Kernel because of its importance in CNN. With the \texttt{top-k} set to six and the \texttt{result num} set to ten, we carried out ten layers for each configuration of the convolutional kernel width (ranging from two to six) and computed the averaged normalized values of the information entropy of the generated skeletons across all trials. The relationship between the number of convolutional layers and the scores is shown in Figure \ref{fig:conv}, where the experimental results demonstrate that the peak performance occurs at 7 convolutional layers. Additionally, Figure \ref{fig:recep} illustrates that the value reaches its maximum when the width is 3 at layer 7. This observation is in accordance with the theoretical design principle: A lack of sufficient layers and a narrow width are unable to capture global contextual information, whereas an excessive number of layers and an overly wide width may lead to the aggregation of redundant information beyond the model's processing capability.



\paragraph{Best-of-N Self-Refinement}

We question whether simply scaling the number of self-refinements can bring continuous improvements. With convolution-related hyperparameters fixed and \texttt{best-of} set to 3, we test and record the information entropy in each self-refined skeleton. As shown in Figure \ref{fig:self}, the peak performance is attained at three self-refined iterations. We can conclude that moderate self-refinement can enhance quality, while excessive self-refinement may lead to deviation from the original material which will cause the deterioration of the skeleton.


\section{Related Work}
Currently, long-to-long generation methods predominantly rely on extractive approaches. For instance, STORM and Co-STORM \citep{storm, costorm} utilize a multi-agent system to formulate questions from diverse perspectives, enabling the retrieval of documents from the Internet for the purpose of authoring a Wiki article. OmniThink \citep{omnithink} and the IRP framework \citep{balepur-etal-2023-expository} enhance the RAG-based method by extracting relevant paragraphs for content writing. Existing end-to-end generation works, such as \citep{bai2025longwriter}, due to the limitations of the model's capabilities, achieving satisfactory results remains challenging.  

Specifically, within the domain of survey writing, \citet{autosurvey} put forward AutoSurvey, a system engineered to automate the process of survey creation via retrieval and iterative refinement. \citet{hu2024hireviewhierarchicaltaxonomydrivenautomatic} presented HiReview, which hierarchically clusters paper titles to generate a skeleton used to produce the full survey content. PaSa \citep{he2025pasallmagentcomprehensive} provide an advanced Paper Search agent. In the current scenario, the consideration of how to integrate vast amounts of information has become increasingly crucial.

\section{Conclusion}

We introduce LLM$\times$MapReduce-V2, an integrative framework that leverages entropy-driven convolutional test-time scaling to enhance the ability of LLMs to process and synthesize extremely long input materials. For evaluation, we present SurveyEval, a novel benchmark designed to assess the effectiveness of our method, demonstrating its superiority over existing baselines in generating comprehensive surveys. Our work contributes both to the theoretical understanding and the technical advancements in long-to-long, resource-intensive generation tasks.

\section*{Limitations}
At present, LLM$\times$MapReduce-V2 has only been verified on the survey task, and in the future, it needs to be extended to more practical tasks, such as research reports.
Benefiting from the high cost-effectiveness and high response speed of the Gemini-flash-thinking model, we mainly conducted experiments based on this model. In the future, we will verify the effectiveness of the method on newer and more powerful models, such as DeepSeek-R1.
The hallucination of the base model may lead to errors and misleading information in the generated, readers need to distinguish the authenticity of the content.

\bibliography{custom}

\appendix
\clearpage

\section{Information Bottleneck in Survey Generation}
\label{sec:ib_appendix}

Let $X$ be the input materials, which include the topic $T$ of the required output article (i.e., $Y$) and the provided resources $R$, which may be very lengthy. For intermediate representations, we introduce the skeleton $S$, aligned with the output $Y$, and the digests $D$, which are compressed summaries derived from the resources $R$. $H(\cdot)$ represents the information entropy, $I(\cdot, \cdot)$ represents the mutual information. 

Eq. \eqref{eq:base} can be deduced as follow:
\begin{align}
    & IB(X,Y) = I(Z;Y) - \beta I(X;Z), &\\
    & I(Z;Y) = I(D,S;Y), \label{eq:2} &\\
    & I(X;Z) = I(R,T;D,S). \label{eq:3} & 
\end{align}

\eqref{eq:2} is simplified as follows:
\begin{align}
    & I(D,S;Y) = I(S;Y) + I(D;Y|S).  & \label{eq:4} 
\end{align}

\eqref{eq:3} is simplified as follows:
\begin{align}
    & I(R,T;D,S) = I(T;D,S) + I(R;D,S|T), \notag \\
    & I(T;D,S) = I(S;T)+ I(D;T|S), \label{eq:5} \\
    & I(R;D,S|T) =I(S;R|T) + I(D;R|T,S). \label{eq:6}
\end{align}

Assume that reference papers $R$ include all information of Survey Skeleton $S$ and Paper Digests $D$, Survey Skeleton $S$ and Paper Digests $D$ include all information of Survey Topic $T$, and Survey $Y$ include all information of Survey Skeleton $S$.

\eqref{eq:4} is simplified as follows:
\begin{align}
    & I(S,Y) = H(S) \label{eq:7} &\\
    & I(D;Y|S) = H(Y|S) - H(Y|D,S) \label{eq:9} &\\
    & H(Y|S) = H(Y) - H(S) \notag& \\
    & H(Y|D,S) = H(Y|D)  &\notag
\end{align}

As $I(Y;D) = H(Y) - H(Y|D)$, so \eqref{eq:9} can be simplified as follow:
\begin{align}
    & I(D;Y|S) = I(Y;D) - H(S) \label{eq:8} \quad \quad \quad & 
\end{align}

Add \eqref{eq:7} and \eqref{eq:8}, \eqref{eq:2} and \eqref{eq:4} can be simplified as:
\begin{align}
    & I(Z;Y)= I(D,S;Y) = I(Y,D) \label{eq:13} \quad \quad&
\end{align}

\eqref{eq:5} can be simplified as:
\begin{align}
    & I(T;D,S) = I(S;T)+ I(D;T|S) \notag \quad \quad \quad&\\ 
    & I(S;T) = H(T) \notag &\\
    & I(D;T|S) = H(T|S) = 0 \notag &\\
    & I(T;D,S) = H(T) \label{eq:10} &
\end{align}

\eqref{eq:6} can be simplified as:
\begin{align}
    & I(R;D,S|T) =I(S;R|T) + I(D;R|T,S) \notag \\
    & I(S;R|T) = H(S|T) = H(S)-H(T) \notag \\
    & I(D;R|T,S) = H(D|T,S) \notag \\
    & H(D|T,S) = H(D) - I(D;T,S) \notag \\
    & I(D;T,S) = H(T,S) = H(S) \notag \\
    & I(R;D,S|T) = H(D) - H(T) \label{eq:11}
\end{align}

Add \eqref{eq:10} and \eqref{eq:11}, Formula \eqref{eq:3} can be simplified as follows
\begin{align}
     &I(X;Z) = H(D) \quad \quad \quad \quad \quad \quad \quad&
\end{align}

Add \eqref{eq:13} and \eqref{eq:12},, we get the result: 
\begin{align}
    & IB(X,Y) = I(Y,D) - \beta H(D) \label{eq:12} \quad \quad \quad
\end{align}

Based on assumptions, we can get this result:
\begin{align}
    & \mathrm{min}(I(Y,D), H(S))\le I(Y,D) \le H(Y,D) 
\end{align}

It can be concluded the upper and lower bounds of IB, namely: 
\begin{align}
\begin{split}
    IB(X,Y) &\ge \mathrm{min}((1-\beta)H(D) - H(D|Y) \\
    &, H(S)-\beta H(D)), \\ 
    IB(X,Y) &\le H(Y|D) + (1-\beta)H(D).
\end{split}
\end{align}

\begin{table*}[t]
\small
\centering
\begin{adjustbox}{width=\textwidth}
\begin{tabular}{lccccccccccc}
\toprule
\multirow{2}{*}{\bf Name} & \multicolumn{5}{c}{\bf Component} & \multicolumn{2}{c}{\bf Input Length} & \multicolumn{2}{c}{\bf Data Size} &\\
\cmidrule(lr){2-6} \cmidrule(lr){7-8} \cmidrule(lr){9-10}
& \multirow{2}{*} {\bf Outline} & \multirow{2}{*} {\bf Abstract} & \multirow{2}{*}{\bf Full Text} & \multicolumn{2}{c}{\bf Refs} & \multirow{2}{*}{\bf Sents. / Token}  & \multirow{2}{*} {\bf Avg. Ref.} & \multirow{2}{*}{\bf Survey Num.} & \multirow{2}{*} {\bf Ref Num.} & \\
\cmidrule(lr){5-6}
& & & & \bf Abstract & \bf Full Text & & & & & & \\
\midrule
AutoSurvey ~\cite{autosurvey}&\textcolor{red}{\ding{55}} & \textcolor{DarkGreen}{\ding{51}}& \textcolor{DarkGreen}{\ding{51}}&\textcolor{red}{\ding{55}} &\textcolor{red}{\ding{55}} &\textendash  &\textendash &\textendash  &530,000 \\
HiCaD ~\cite{hu2024hireviewhierarchicaltaxonomydrivenautomatic} & \textcolor{DarkGreen}{\ding{51}} &\textcolor{red}{\ding{55}} &\textcolor{red}{\ding{55}} &\textcolor{DarkGreen}{\ding{51}} &\textcolor{red}{\ding{55}} & 471.4 / \textendash &81.1 &7,637 & 619,360 \\
NLPCC2024 Shared Task 6 ~\cite{10.1007/978-981-97-9443-0_35}&\textcolor{red}{\ding{55}} &\textcolor{DarkGreen}{\ding{51}} &\textcolor{DarkGreen}{\ding{51}} & \textcolor{DarkGreen}{\ding{51}}&\textcolor{red}{\ding{55}} &\textendash  &98.5 & 700 & 68,950\\
SciReviewGen ~\cite{kasanishi-etal-2023-scireviewgen} & \textcolor{DarkGreen}{\ding{51}}& \textcolor{DarkGreen}{\ding{51}}& \textcolor{DarkGreen}{\ding{51}}& \textcolor{DarkGreen}{\ding{51}}&\textcolor{red}{\ding{55}}  & \textendash  { }/ 12.5k &68 &10,130 & 690,000 \\
BigSurvey ~\cite{ijcai2022p591}&\textcolor{red}{\ding{55}} & \textcolor{DarkGreen}{\ding{51}}& \textcolor{DarkGreen}{\ding{51}}& \textcolor{DarkGreen}{\ding{51}}&\textcolor{red}{\ding{55}} &450.1 / \textendash &76.3 &4,478 & 341,671 \\
SurveySum ~\cite{fernandes2024surveysumdatasetsummarizingmultiple} & \textcolor{DarkGreen}{\ding{51}}&\textcolor{red}{\ding{55}} & \textcolor{DarkGreen}{\ding{51}}&\textcolor{red}{\ding{55}} & \textcolor{DarkGreen}{\ding{51}}&\textendash &\textendash & 6 & \textendash \\
\cmidrule(lr){1-11}
\textbf{SurveyEval} & \textcolor{DarkGreen}{\ding{51}}& \textcolor{DarkGreen}{\ding{51}}& \textcolor{DarkGreen}{\ding{51}}& \textcolor{DarkGreen}{\ding{51}}& \textcolor{DarkGreen}{\ding{51}}&27.5k / 1383.2k &110.6 &384 & 42,480 \\
\textbf{SurveyEval-test} & \textcolor{DarkGreen}{\ding{51}}& \textcolor{DarkGreen}{\ding{51}}& \textcolor{DarkGreen}{\ding{51}}& \textcolor{DarkGreen}{\ding{51}}& \textcolor{DarkGreen}{\ding{51}}&40.8k / 2112.0k &179.3 &20 & 3,585\\
\bottomrule
\end{tabular}
\end{adjustbox}
\caption{Comparison of survey datasets, highlighting key components, input lengths, and data sizes across multiple datasets. The \textit{Component} column shows the inclusion of specific parts in each dataset: \textit{Outline}, \textit{Abstract}, \textit{Full Text}, and references (\textit{Refs}), with the \textit{Refs} column further split into references in the Abstract and Full Text. The \textit{Input Length} section provides the average number of sentences (Sents.) and tokens (Token) per data entry, while \textit{Avg. Ref.} denotes the average number of references per entry. The \textit{Survey Num.} indicates the number of surveys included in the dataset, and \textit{Ref. Num.} reflects the total number of references for the surveys. For datasets without publicly available information, a "\textendash" is used as a placeholder.}
\label{tab:Comparison of datasets}
\end{table*}

\section{Details of SurveyEval Dataset}
\label{sec:Detail_Test_Surveys}
\subsection{Dataset Construction}
The limitations of currently available publicly released survey datasets are evident, as they predominantly include only abstracts of the references, which often lack the detailed information necessary for comprehensive survey-based research. For instance, the \textit{AutoSurvey} dataset~\cite{autosurvey} does not include any reference relationships, while others, such as \textit{HiCaD}~\cite{hu2024hireviewhierarchicaltaxonomydrivenautomatic}, focus primarily on the outlines of literature surveys. Additionally, datasets like \textit{NLPCC2024 Shared Task 6}~\cite{10.1007/978-981-97-9443-0_35}, \textit{SciReviewGen}~\cite{kasanishi-etal-2023-scireviewgen}, and \textit{BigSurvey}~\cite{ijcai2022p591} only include abstracts of references, which limits their applicability for more in-depth research tasks.

Moreover, the few datasets that do include full-text references are generally tailored to section generation tasks, and the \textit{SurveySum} dataset~\cite{fernandes2024surveysumdatasetsummarizingmultiple} contains only six literature surveys. To bridge this gap and significantly enhance existing frameworks, we constructed the \textbf{SurveyEval Benchmark}. This dataset is designed to contribute to long-to-long generation tasks, which are essential for advancing models' capabilities to handle long-form texts. The SurveyEval dataset is distinctive in its inclusion of both comprehensive literature reviews and \textbf{full references}, along with its superior handling of input length.

Our dataset construction process was carefully designed to ensure both data quality and relevance. We obtained academic survey papers by querying the arXiv repository within the \texttt{cs.CL} category. After filtering the papers using large language models (LLMs) to determine their suitability as academic surveys, we conducted further searches for their references in reputable sources such as ACL, NeurIPS, CVPR, and Google Scholar. To process the raw PDF data, we utilized \texttt{MinerU}~\cite{wang2024mineruopensourcesolutionprecise}, an open-source tool developed for the precise extraction of academic content into a structured Markdown format. After data extraction, we employed a two-step quality control process: (1) automated filtering using the \texttt{Qwen2.5-72B-Instruct-AWQ-YARN-128k} model~\cite{qwen2.5} to remove low-quality papers, and (2) manual verification to ensure the accuracy and relevance of the content.

For a detailed comparison of the dataset characteristics, refer to Table~\ref{tab:Comparison of datasets}.

\subsection{Test Dataset}
Generating, evaluating, and manually assessing survey-based algorithms is a time-consuming and resource-intensive process. Given this, the \textit{AutoSurvey} model~\cite{autosurvey} also uses a test set of 20 papers. Similarly, for this study, we selected 20 papers from the SurveyEval dataset to conduct our research.

To ensure a fair and comprehensive evaluation, we applied two main selection criteria: (1) the completeness of reference retrieval (i.e., the percentage of references successfully obtained from external sources), and (2) the diversity of token counts in the reference lists, ensuring a wide range of input sizes. This approach ensures that our test set is representative of real-world scenarios. Specific details of the dataset can be found in Table~\ref{survey-test-stats}.
\begin{table*}
  \centering
  \begin{adjustbox}{width=\textwidth}

    \footnotesize
\begin{tabular}{lcccc}
    \hline
    \textbf{Title} & \textbf{Survey Token} & \textbf{Ref. Rate} & \textbf{Ref. Count} & \textbf{Ref. Token} \\
    \hline
    Recent Advances in Direct Speech-to-text Translation & 7327 & 100.00\% & 23 & 236824 \\
    A Primer on Contrastive Pretraining in Language Processing: Methods, Lessons Learned and Perspectives & 8367 & 100.00\% & 40 & 495758 \\
    End-to-end Task-oriented Dialogue: A Survey of Tasks, Methods, and Future Directions & 12385 & 100.00\% & 52 & 689330 \\
    A Survey on Proactive Dialogue Systems: Problems, Methods, and Prospects & 8278 & 100.00\% & 56 & 823666 \\
    Modern Question Answering Datasets and Benchmarks: A Survey & 10240 & 100.00\% & 75 & 1294011 \\
    A Survey on Measuring and Mitigating Reasoning Shortcuts in Machine Reading Comprehension & 11099 & 100.00\% & 106 & 2058171 \\
    A Survey on Recent Advances in Reinforcement Learning for Dialogue Policy Learning & 10068 & 99.07\% & 107 & 2123869 \\
    A Survey on Explainability in Machine Reading Comprehension & 9732 & 98.44\% & 125 & 2069256 \\
    Confidence Estimation and Calibration in Large Language Models: A Survey & 11311 & 99.25\% & 128 & 2421195 \\
    Controllable Text Generation with Transformer-based PLMs: A Survey & 20350 & 98.84\% & 170 & 2486701 \\
    Measure and Improve Robustness in NLP Models: A Survey & 9548 & 98.33\% & 177 & 3257176 \\
    Neural Entity Linking: A Survey of Deep Learning Models & 35275 & 98.10\% & 206 & 3373014 \\
    Machine Reading Comprehension: Contextualized Language Models and Beyond & 33695 & 96.77\% & 207 & 4663897 \\
    Non-Autoregressive Generation for Neural Machine Translation: A Survey & 37197 & 97.93\% & 236 & 4254491 \\
    Chain of Thought Reasoning: Advances, Frontiers and Future & 18302 & 95.40\% & 248 & 3233452 \\
    Bias and Fairness in Large Language Models: A Survey & 47372 & 95.59\% & 260 & 677128 \\
    Efficient Methods for Natural Language Processing: A Survey & 12253 & 98.94\% & 280 & 1119131 \\
    The Efficiency Spectrum of Large Language Models: An Algorithmic Survey & 19574 & 94.80\% & 327 & 2128935 \\
    Pre-trained Language Models in Biomedical Domain: A Systematic Survey & 41887 & 95.76\% & 351 & 1426231 \\
    Code-Switching Research in NLP: A Systematic Survey on Trends and Challenges & 13239 & 93.76\% & 411 & 3408349 \\
    \hline
  \end{tabular}
  \end{adjustbox}
  \caption{\label{survey-test-stats}Test Set Statistics of SurveyEval. The \textit{Survey Token} represents the total length of the literature survey in tokens. The \textit{Ref. Rate} indicates the percentage of references that were successfully retrieved and converted into usable data. The \textit{Ref. Count} refers to the total number of references cited in each literature survey. The \textit{Ref. Token} represents the cumulative token count of all references associated with the literature survey. }
  
  \normalsize  
\end{table*}

\section{Implementation Details of Baselines}
\label{sec:baseline_detail}
\subsection{Implementation of Vanilla}
\label{ref:Implementation of Vanilla}
The vanilla baseline serves as a straightforward approach to literature review generation. This implementation makes direct use of the language model’s capabilities by feeding it the survey topic along with the full content of all referenced papers. To address the model’s context window limitations while ensuring comprehensive coverage, we apply a proportional text cropping strategy to the reference papers. For example, we have 3 reference paper, their length are $\alpha$, $\beta$ and $\gamma$ respectively, and the total window size is $W$. In this setting, each paper needs to be cut down to $\frac{W}{\alpha + \beta + \gamma}$.

\subsection{Implementation of Vanilla with skeleton}
\label{ref:Implementation of Vanilla+S}
This baseline improves the survey generation process by adopting a two-stage approach. In the first stage, the model generates a structural skeleton based on the topic and abstracts of all referenced papers. In the second stage, this skeleton is combined with the full text of the referenced articles to produce a comprehensive survey.

\subsection{Implementation of AutoSurvey} \label{sec:AutoSurvey_Details}
In this study, we implement AutoSurvey using the test set from SurveyEval (test set details are provided in Appendix \ref{sec:Detail_Test_Surveys}). We follow the original framework while making necessary adjustments to accommodate our testing dataset and evaluation process. All parameter settings are consistent with those specified in the original work.

\paragraph{Data Adaptation.} To ensure compatibility with our evaluation framework and dataset, we construct a retrieval database for each survey paper and its corresponding references. Although the number of references in our dataset is fewer than 1,200, we still configure the retrieval to include 1,200 papers to ensure all references can be retrieved, as specified in AutoSurvey. This retrieval process ensures that all references for a survey paper are included in the initial retrieval stage.

\paragraph{Subsection and Outline Generation.} 
 The embedding model is \texttt{nomic-embed-text-v1}, in line with the original AutoSurvey implementation. All parameters remain unchanged from the original paper. Outline generation is based on the abstracts of the selected papers, as in the original method. For subsection generation, the number of sections is predetermined to be 8. The model processes the first 1,500 tokens from the main body of the 60 relevant papers retrieved, ensuring detailed and coherent descriptions. The same set of reference papers is used throughout the reflection and polishing stages to maintain consistency and accuracy.

\subsection{Implementation of LLM\texorpdfstring{$\times$}{x}MapReduce-V2} 
\label{sec:LLMxMapReduce-V2_Details}
Here are the important hyperparameters of LLM$\times$MapReduce-V2:
\begin{itemize}
    \item convolution\_layer = 6, as Digest-Based Feedback Clustering equals to one layer
    \item kernel\_width = 3
    \item convolution\_result \_num = 10
    \item top\_k = 6
    \item self\_refine\_count = 3
    \item self\_refine\_best\_of = 3
\end{itemize}

\section{Details of SurveyEval Evaluation}
\begin{figure*}[ht]
    \centering
    \includegraphics[width=1\linewidth]{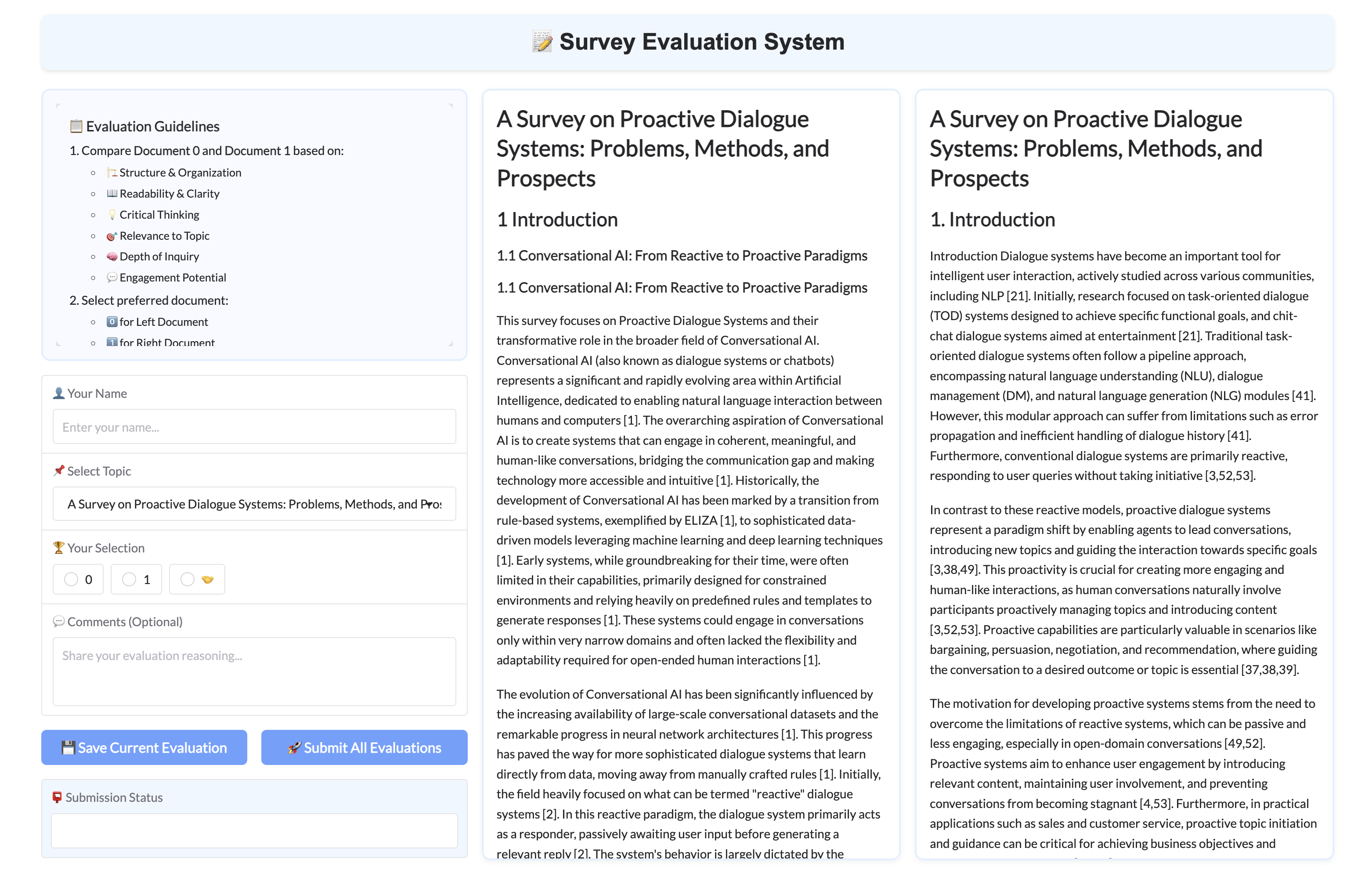}
    \caption{Screenshot of the web application for evaluating the survey pair.}
    \label{fig:human—eval}
\end{figure*}

Our evaluation framework consists of both automatic and human evaluation components to ensure a comprehensive assessment. To standardize the evaluation across multiple dimensions, we set the score range for all assessments to a 100-point scale. To facilitate an objective comparison with the baseline, we have referenced specific evaluation metrics from AutoSurvey~\cite{autosurvey}. The original automated evaluation metrics include two main components: Content Quality and Citation Quality.

For Content Quality, we retained the criteria of structure and relevance. Since the original scoring used a 5-point scale, we multiplied the raw scores by 20 after obtaining them to enhance differentiation and align the scores with other ranges.

The original coverage score has been refined and is now represented by a more detailed assessment of reference quality. 
As for Citation Quality, we adapted the evaluation prompt from AutoSurvey~\cite{autosurvey} for individual citations while modifying and supplementing the calculation methods. Below are the specific criteria and the implementation of the SurveyEval Evaluation:

\subsection{Automatic Evaluation criteria}
\label{ref:SurveyEval criteria}
\subsubsection{Structure Quality criteria}
\label{sec:structure_detail}
The structure of the survey is evaluated based on the criteria outlined in AutoSurvey. The score, initially on a scale of 0-5, is multiplied by 20 to align with other score ranges.  For the detailed criteria, please refer to Table~\ref{tab: structure}.

\begin{table*}[ht]
\centering
\begin{tabular}{|c|>{\raggedright\arraybackslash}p{10cm}|}
\hline
\textbf{Description} & \small Structure: Structure evaluates the logical organization and coherence of sections and subsections, ensuring that they are logically connected. \\
\hline
\textbf{Score 1} & \small The survey lacks logic, with no clear connections between sections, making it difficult to understand the overall framework. \\
\hline
\textbf{Score 2} & \small The survey has weak logical flow with some content arranged in a disordered or unreasonable manner. \\
\hline
\textbf{Score 3} & \small The survey has a generally reasonable logical structure, with most content arranged orderly, though some links and transitions could be improved such as repeated subsections. \\
\hline
\textbf{Score 4} & \small The survey has good logical consistency, with content well arranged and natural transitions, only slightly rigid in a few parts. \\
\hline
\textbf{Score 5} & \small The survey is tightly structured and logically clear, with all sections and content arranged most reasonably, and transitions between adjacent sections smooth without redundancy. \\
\hline
\end{tabular}
\caption{Structure Evaluation Criteria}
\label{tab: structure}
\end{table*}

\subsubsection{Content Quality criteria}
\label{sec:content_detail}
For \textbf{Faithfulness}, we adopted the prompt from AutoSurvey ~\cite{autosurvey} shown in Fig.~\ref{fig:NLI-prompt} to assess citation quality, evaluating the accuracy and relevance of citations within the survey. For the CLAIM component, we mapped citation indices to their corresponding reference papers, conducting separate evaluations for multiple citations to ensure each assessment is associated with only one reference paper. For the SOURCE component, we incorporated the full text of the corresponding reference paper. The detailed Faithfulness is calculated as follows:
\begin{equation}
\scalebox{1}{$\displaystyle
\text{Faithfulness} = \frac{\sum_{i=1}^{C} \mathbb{I} \left[\sum_{j=1}^{R_{c_i}} h(c_i, r_j) \right]}{C},$}
\nonumber
\end{equation}  
where $R_{c_i}$ is the number of times paper $c_i$ is cited, $C$ is the number of claims in the survey, and $r_j$ represents the $j$th cited reference paper of $c_i$,
\begin{equation} 
    \scalebox{1}{$
        \displaystyle h(c_i, r_j) = \begin{cases}   
            1, & \text{if } r_j \text{ correctly supports } c_i \\   
            0, & \text{otherwise}   
        \end{cases} 
    $} 
    \nonumber 
\end{equation}
, and 
\begin{equation} 
    \scalebox{1}{$
        \displaystyle \mathbb{I}(x) = \begin{cases}   
            1, & \text{if } x \ge 1 \\   
            0, & \text{otherwise}   
        \end{cases} 
    $} 
    \nonumber 
\end{equation}

\begin{figure*}[h]
	\begin{tcolorbox}[colback=blue!2,colframe=blue!50!black]
	\small
	\texttt{{-}{-}{-}\\
Claim:\\
\texttt{[CLAIM]}\\
{-}{-}{-}\\
Source: \\
\texttt{[SOURCE]}\\
{-}{-}{-}\\
Claim:\\
\texttt{[CLAIM]}\\
{-}{-}{-}\\
Is the Claim faithful to the Source? \\
A Claim is faithful to the Source if the core part of the Claim can be supported by the Source.\\\\
Only reply with 'Yes' or 'No':
}
	\end{tcolorbox}
	\caption{Claim evaluation prompt.}
	\label{fig:NLI-prompt}
\end{figure*}

The \textbf{Relevance} of the survey is also evaluated based on the criteria from AutoSurvey ~\cite{autosurvey}. The score, initially on a scale of 0-5, is multiplied by 20 to align with other score ranges. For detailed criteria, please refer to Table~\ref{tab:relevance}.    \begin{table*}[ht]
    \centering
    \begin{tabular}{|c|>{\raggedright\arraybackslash}p{10cm}|}
    \hline
    \textbf{Description} & \small Relevance: Relevance measures how well the content of the survey aligns with the research topic and maintains a clear focus. \\
    \hline
    \textbf{Score 1} & \small The content is outdated or unrelated to the field it purports to review, offering no alignment with the topic. \\
    \hline
    \textbf{Score 2} & \small The survey is somewhat on topic but with several digressions; the core subject is evident but not consistently adhered to. \\
    \hline
    \textbf{Score 3} & \small The survey is generally on topic, despite a few unrelated details. \\
    \hline
    \textbf{Score 4} & \small The survey is mostly on topic and focused; the narrative has a consistent relevance to the core subject with infrequent digressions. \\
    \hline
    \textbf{Score 5} & \small The survey is exceptionally focused and entirely on the topic; the article is tightly centred on the subject, with every piece of information contributing to a comprehensive understanding of the topic. \\
    \hline
    \end{tabular}
    \caption{Relevance Evaluation Criteria}
    \label{tab:relevance}
    \end{table*}

To provide a more comprehensive evaluation of the quality of the generated literature reviews, we propose two additional evaluation criteria: \textbf{Language} and \textbf{Criticalness}. Language evaluates the clarity, formality, and redundancy in the writing, ensuring it maintains academic rigour while avoiding unnecessary repetition. Criticalness assesses the depth of analysis, originality of insights, and the identification of future research directions. For detailed scoring standards, please refer to Figure~\ref{fig:language-prompt} and Figure~\ref{fig:critical-evaluation-prompt}.

\begin{figure*}[h]
	\begin{tcolorbox}[colback=blue!2,colframe=blue!50!black]
	\small
	\texttt{\texttt{[Task]}\\
Rigorously evaluate the quality of an academic survey about \texttt{[TOPIC]} by scoring three dimensions (each 0-100) and calculating the average as the final score. \\\\
\texttt{[Evaluation Criteria]}\\
Evaluate each dimension on a 0-100 scale based strictly on the highest standards below. The final score is the average of the three dimension scores.\\\\
1. **Academic Formality** (100 points):  \\
   - Demonstrates *flawless* academic rigor. Uses precise terminology consistently, avoids colloquial language entirely, and maintains a strictly scholarly tone. Sentence structures are sophisticated and purposefully crafted to enhance analytical depth. **Even a single instance of informal phrasing or imprecise terminology disqualifies a perfect score**.\\\\
2. **Clarity \& Readability** (100 points):  \\
   - Writing is *exceptionally* clear and concise. Sentences are logically structured, with no ambiguity. Transitions between ideas are seamless, and the argument progresses with precision. **Any unnecessary complexity or minor ambiguity precludes full marks.**  \\\\
3. **Redundancy** (100 points):  \\
   - **Unique**: each sentence must have a unique value and cannot be repeated. Repetition is only allowed to maintain structural coherence, such as using uniform terminology or necessary transitional phrases. Repeating key concept definitions in a new context to help readers understand can be seen as a structural requirement.\\
   - **Efficient argumentation**: Argumentation needs to be efficient, with logically coherent viewpoints and avoiding unnecessary repetition. Even minor repetitions without actual structural effects can result in the deduction of points. For example, repeating a discovery almost identical in the same paragraph without providing new insights or perspectives will result in the deduction of points. \\\\
\texttt{[Topic]}\\
\texttt{[TOPIC]}\\\\
\texttt{[Section]}\\
\texttt{[SECTION]}\\\\
\texttt{[Output Format]}\\
Rationale:\\
\textless Provide a detailed reason for the score, considering all dimensions step by step. Highlight specific strengths and weaknesses, such as the consistency of academic tone, the clarity of sentence structure, or the presence of redundancy.\textgreater \\
Final Score: \\
\textless SCORE\textgreater (X+Y+Z)/3\textless/SCORE\textgreater  \\
(Example: \textless SCORE\textgreater23\textless/SCORE\textgreater; scores can include two decimal place)
}
	\end{tcolorbox}
	\caption{Language evaluation prompt.}
	\label{fig:language-prompt}
\end{figure*}
\begin{figure*}[h]
	\begin{tcolorbox}[colback=blue!2,colframe=blue!50!black]
	\small
	\texttt{\texttt{[Task]}\\
Rigorously evaluate the quality of an academic survey about \texttt{[TOPIC]} by scoring three dimensions (each 0-100) and calculating the average as the final score. \\\\
\texttt{[Evaluation Criteria]}\\
The final score is the sum of the individual scores from the following three dimensions. Please evaluate each dimension thoroughly and rigorously.\\\\
1. **Critical Analysis** (100 points): \\
   - Offers a deep, incisive critique of methodologies, results, and underlying assumptions. Provides a clear identification of significant gaps, weaknesses, and areas for improvement. Challenges assumptions with well-supported arguments, offering clear alternatives or improvements. \\\\
2. **Original Insights** (100 points): \\
   - Proposes novel, well-supported interpretations or frameworks based on the reviewed literature. Demonstrates a strong understanding of the subject matter and provides genuinely original contributions that challenge the status quo. Insights are clearly connected to existing research, offering fresh perspectives or unique ways forward. \\\\
3. **Future Directions** (100 points): \\
   - Clearly identifies specific, promising research directions with strong justification. Suggests actionable, concrete ideas for future research that are rooted in the gaps identified within the reviewed literature. Demonstrates foresight in proposing innovative approaches and methodologies. \\\\
\texttt{[Topic]}\\
\texttt{[TOPIC]}\\\\
\texttt{[Section]}\\
\texttt{[SECTION]}\\\\
\texttt{[Output Format]}\\
Rationale:\\
\textless Provide a detailed reason for the score, considering all dimensions step by step. Highlight specific strengths and weaknesses, such as the depth of critique, the originality of insights, or the clarity of future directions.\textgreater \\
Final Score: \\
\textless SCORE\textgreater (X+Y+Z)/3\textless/SCORE\textgreater  \\
(Example: \textless SCORE\textgreater23\textless/SCORE\textgreater; scores can include two decimal places)
}
	\end{tcolorbox}
	\caption{Criticalness evaluation prompt.}
	\label{fig:critical-evaluation-prompt}
\end{figure*}

\subsubsection{Claim Evaluation Details}
\label{sec:SurveyEval CD}

\paragraph{Claim Numbers}
Inspired by FactScore's approach to decomposing atomic knowledge~\cite{factscore}, we adapt its methodology to extract effective claims from the paper. Specifically, each section of the survey is treated as an independent unit, with claims extracted separately for each. The extraction process employs a structured, prompt-based approach using the \texttt{gemini-2.0-flash-thinking-exp-1219} model, which adheres to specific consolidation rules for claim identification. The extraction prompt enforces strict guidelines, as shown in Fig.~\ref{fig:claim-decomposition-prompt}.

\begin{figure*}[h]
	\begin{tcolorbox}[colback=blue!2,colframe=blue!50!black]
	\small
\texttt{Analyze the following text and decompose it into independent claims following strict consolidation rules:
 \\\\ \texttt{[Claim Definition]}\\
A verifiable objective factual statement that functions as an independent knowledge unit. Each claim must: \\
1. Contain complete subject-predicate-object structure\\
2. Exist independently without contextual dependency\\
3. Exclude subjective evaluations \\\\
\texttt{[Merge Rules]}$\rightarrow$ Should merge when: \\
- Same subject + same predicate + different objects (e.g., "Should measure A / Should measure B" $\rightarrow$ "Should measure A and B") \\
- Different expressions of the same research conclusion \\
- Parallel elements of the same category (e.g., "A, B and C") \\\\
\texttt{[Separation Rules]}$\rightarrow$ Should keep separate when: \\
- Different research subjects/objects \\
- Claims with causal/conditional relationships \\
- Findings across temporal sequences \\
- Conclusions using different verification methods \\\\
\texttt{[Output Format]}\\
Strict numbered list with consolidated claims maintaining grammatical integrity: \\
1. Use "and/or/including" for merged items \\
2. Separate parallel elements with commas \\
3. Prohibit abbreviations or contextual references \\\\
Below is the text you need to extract claims from: \\\\
\texttt{\{text\}}\\\\
}
	\end{tcolorbox}
	\caption{Claim decomposition prompt.}
	\label{fig:claim-decomposition-prompt}
\end{figure*}

To ensure uniqueness, we implement a two-phase deduplication process. The first phase performs intra-group deduplication on smaller batches (300 claims each), while the second phase conducts cross-group deduplication, deduplicates pairwise and thenmergese them until there is only one group left. Both phases utilize the deduplication criteria outlined in Figure~\ref{fig:redundancy-removal-prompt}. The final claim number is determined based on the total number of claims after deduplication.

\begin{figure*}[h]
	\begin{tcolorbox}[colback=blue!2,colframe=blue!50!black]
	\small
	\texttt{Below is a numbered list of claims. Your task is to identify and group claims that convey the same information, removing all redundancy. \\\\ \texttt{[Guidelines]}\\
- Claims that express the same fact or knowledge in different wording or detail are duplicates. \\
- If one claim is fully included within another or repeats the same idea, consider it a duplicate. \\
- Claims with differing details, context, or scope are not duplicates. \\\\
For each group of duplicates, output the serial numbers of the claims to be removed (comma-separated). Choose one claim to keep. \\\\
Example: \\
If claims 2, 5, and 8 are duplicates and claim 2 is kept, output "5,8". \\\\
\texttt{List of claims:}\\
\texttt{\{numbered\_facts\}}\\\\
Output ONLY the serial numbers to remove. No additional text. 
}
	\end{tcolorbox}
	\caption{Redundancy removal prompt.}
	\label{fig:redundancy-removal-prompt}
\end{figure*}

\paragraph{Claim Density.}
Claim Density is defined as the ratio of unique claims to the total number of extracted claims prior to deduplication. This metric serves as a measure of information redundancy in the original text, with a higher density indicating a more efficient presentation of information. The density is computed after both intra-group and cross-group deduplication phases to ensure that only genuinely unique claims are included in the final count. It can be calculated as follows:

\begin{equation}
\text{Claim Density} = \frac{\delta(c_{ij})}{\sum_{i=1}^{S} \sum_{j=1}^{C_i}},
\nonumber
\end{equation}

   where \( C_i main\) represents the number of claims extracted from section \( i \), \( S \) is the total number of sections. \( c_{ij} \) represents the \( j \)th claim in section \( i \) and \( \delta(\cdot) \) is an indicator function that:  
   \begin{equation}
   \delta(\cdot) =
   \begin{cases} 
   1, & \text{if } \cdot \text{ is retained as unique} \\
   0, & \text{if } \cdot \text{ is redundant}
   \end{cases}
   \nonumber
   \end{equation}

\subsubsection{Reference Evaluation Details}
\label{sec:Reference Evaluation Details}
In order to measure the utilization rate of the provided references, two metrics are designed:

\paragraph{Precision} measures the coverage of input references by verifying whether each reference is correctly cited at least once.
It is calculated as:  
\begin{equation}
\scalebox{1.0}{$\displaystyle
\text{Ref. P} = \frac{\sum_{j=1}^{R} \mathbb{I} \left[\sum_{i=1}^{C} h(c_i, r_j) \right]}{R}$},
\nonumber
   \end{equation}  
where $ R $ is the number of input references, $ C $ is the number of sentences with citations in the survey, $r_j$ is the $j$th reference paper and  
\begin{equation}
\scalebox{1.0}{$\displaystyle
h(c_i, r_j) = \begin{cases}  
1, & \text{if } r_j \text{ correctly supports } c_i \\  
0, & \text{otherwise}  
\end{cases} $}
\nonumber
   \end{equation}
\paragraph{Recall} evaluates the total number of input references that appear at least once in the generated survey.
It is calculated as:  
\begin{equation}
\text{Ref. R} = \frac{\sum_{i=1}^{R} c(r_i)}{R},
\nonumber
 \end{equation}
where  
\begin{equation}
c(r_i) = \begin{cases}  
1, & \text{if } r_i \in R_S \\  
0, & \text{otherwise}  
\end{cases}
\nonumber
 \end{equation}
and $ R_S $ denotes the set of references appearing in the survey.

\subsection{Human Evaluation Details}
\label{sec:Human Evaluation Details}


The evaluation process was designed to ensure randomization of topics in order to minimize any potential bias. Evaluators were instructed to select their preferred survey by choosing either "Document 0," "Document 1," or marking "Tie" if both documents were of equal quality. Additionally, evaluators were encouraged to provide comments explaining their choices.

Figure \ref{fig:human—eval} shows a screenshot of the evaluation interface. All results were recorded in real-time and saved for subsequent analysis.

We selected 20 topics from the test set, which were consistent with those used in the automatic evaluation. A total of 17 volunteers from the university were recruited, resulting in 217 valid data points, with the win rate displayed in Figure \ref{fig:win_rate}.

\end{document}